\documentclass[conference]{IEEEtran}
\usepackage{cite}
\usepackage{amsmath,amssymb,amsfonts}
\usepackage{algorithmic}
\usepackage{graphicx}
\usepackage{textcomp}
\usepackage{xcolor}
\usepackage{subfig}
\usepackage{url}

\def\BibTeX{{\rm B\kern-.05em{\sc i\kern-.025em b}\kern-.08em
    T\kern-.1667em\lower.7ex\hbox{E}\kern-.125emX}}
\begin{document}

\title{SimuShips - A High Resolution Simulation Dataset for Ship Detection with Precise Annotations \\
}
\author{\IEEEauthorblockN{Minahil Raza,
Hanna Prokopova, Samir Huseynzade, Sepinoud Azimi and
Sebastien Lafond}
\IEEEauthorblockA{Faculty of Science and Engineering,
Åbo Akademi University\\
Turku, Finland\\
name.surname@abo.fi
}}


\maketitle

\begin{abstract}
Obstacle detection is a fundamental capability of an autonomous maritime surface vessel (AMSV). State-of-the-art obstacle detection algorithms are based on convolutional neural networks (CNNs). While CNNs provide higher detection accuracy and fast detection speed, they require enormous amounts of data for their training. In particular, the availability of domain-specific datasets is a challenge for obstacle detection. The difficulty in conducting onsite experiments limits the collection of maritime datasets. Owing to the logistic cost of conducting on-site operations, simulation tools provide a safe and cost-efficient alternative for data collection. In this work, we introduce SimuShips, a publicly available simulation-based dataset for maritime environments. Our dataset consists of 9471 high-resolution (1920x1080) images which include a wide range of obstacle types, atmospheric and illumination conditions along with occlusion, scale and visible proportion variations. We provide annotations in the form of bounding boxes. In addition, we conduct experiments with YOLOv5 to test the viability of simulation data. Our experiments indicate that the combination of real and simulated images improves the recall for all classes by 2.9\%.
\end{abstract}


\section{Introduction}
Maritime transportation is a key part of the international supply chain along with an important means of transportation. According to forecasts, the global shipping traffic is expected to grow by 4\% each year~\cite{wwfmmi}. This increasing demand for water-based navigation has created an opportunity for autonomous maritime surface vessels (AMSVs). The interest in AMSVs is shared by both- academia and industry. In particular, several projects focus on collision avoidance and path planning for autonomous navigation of AMSVs~\cite{munin, autonomousshipproject, roboat}. The key motivator for unmanned operations is to reduce human error which is major causal factor in accidents at the sea. According to the European Maritime Safety Agency, a total of 2632 accidents causing 36 deaths and 618 injuries were reported in 2020 alone\cite{emsa}. Out of these accidents, around 43\% were caused by navigational events such as collisions, grounding, and contact. Human-related factors and operational issues were found to be the major causes behind these accidents. Therefore, the development of AMSVs can help in automating operations leading to a reduction in such navigational events.

The first step toward autonomous operations is to create situational awareness about the environment. Previously, this task was extensively dependent on human monitoring making it mundane and unproductive. On the contrary, the same task can be performed by equipping the vessel with sensors such as cameras, LiDAR, GNSS and IMU. Recent works on autonomous vessels deploy machine learning and deep learning to process data from onboard sensors~\cite{noel_a_2019_3572368}. Particularly, the data collected from cameras has attracted attention in the field of navigation. This data is mainly used for the detection of obstacles for collision avoidance. Therefore, object detection is a core part of maritime intelligent ship safety technology \cite{yolov4_maritime, yolov5_underwater}. 

Given the complex nature of marine environments, traditional computer vision methods such as background subtraction and image segmentation fail to capture the objects under the influence of sunlight, waves, and weather conditions. On the contrary, deep learning-based algorithms are more robust and accurate in detecting objects. These algorithms have a high potential to navigate ships safely by identifying and tracking obstacles at sea. In particular, convolutional neural networks (CNNs) have attracted a lot of attention in recent years for the object detection task. State-of-the-art object detection algorithms such as YOLO~\cite{yolo}, RCNN~\cite{rcnn}, Faster-RCNN~\cite{faster_rcnn} and SSD~\cite{ssd} are based on CNNs. 

CNNs provide higher detection accuracy and fast detection speed. The structure of CNNs consists of multiple filters spread across layers which are trained to capture different features of an input image. However, these networks consist of millions of parameters. Therefore, they require enormous amounts of data for their training. Due to the scarcity of annotated data, the availability of domain-specific datasets is an essential problem in object detection. Even with the availability of several datasets, the amount of data in the maritime domain is still scarce. Moreover, most of the publicly available maritime datasets \cite{data1, data2, data3} focus on data recorded from the shore creating a shortage of onboard image and video data. The main reason behind this scarcity of data is the difficulty in conducting onsite experiments. Therefore, existing datasets are extremely limited in terms of variety of illumination and weather conditions, number of annotated objects per image and depiction of real scenarios in maritime imagery. Owing to the logistic cost of conducting on-site operations, algorithms are developed in simulation environments such as ROS and Gazebo. Although simulation tools provide a safe and cost-efficient alternative to experimentation, they have not been exploited to collect data or train object detection algorithms. The major limitation is the lack of incorporation of real-life visuals in the simulation environment. 

Therefore, we make the following contributions with this work:
\begin{itemize}
    \item We provide the first publicly available simulation-based dataset for maritime environments consisting of 9471 high-resolution (1920x1080) images resembling the real world with precise annotations
    \item We achieve an improvement of 2.9\% in the recall with YOLOv5~\cite{yolov5docs} through the addition of simulated data proving its benefits 
\end{itemize}

In addition, we provide annotations for object detection along with annotation scripts. Our open-source implementation provides a Jupyter notebook to visualize the dataset and plot bounding boxes for further exploration.

The presented dataset can be retrieved from Zenodo at  \textcolor{blue}{\url{https://doi.org/10.5281/zenodo.7003924}}. The Github repository hosting the source code for data annotation is available at \textcolor{blue}{\url{https://github.com/MinahilRaza/SimuShips-visualizations}}.

The rest of this paper is organized as follows. Section \ref{methodology} presents the methodology for data collection and annotation. Section \ref{exp} discusses the experiments with real and simulated data along with the results. Lastly, Section \ref{conclusion} concludes this work.

\section{Methodology} \label{methodology}
\subsection{Data Collection through AILiveSim}
The dataset was acquired using a simulation tool, AILiveSim \cite{ailivesim}. AILiveSim is a 3D simulation platform for the targeted scalable development and integration of autonomous systems through digital twins. The tool provides a realistic 3D model of the route of the watercraft extended from the city of Turku to Ruisalo in South-West Finland. It features identical physical characteristics and configurations of sensors and ships for data collection. Furthermore, it is possible to add stationary or dynamic ships that can be found sailing the Aura river and Baltic Sea. For data collection, four cameras were placed at a $90^\circ$ angle to each other to capture images from different perspectives. The images were captured at 30 frames per second and stored in FullHD (1920x1080). In addition to the RGB images, the corresponding segmentation masks were extracted as well. While the route remained largely the same, different scenarios and situations were created by placing different sets of objects at different locations. Each situation contained multiple types of ships and obstacles varying in size, colour and motion. The simulation time was approximately 90 minutes. 

\subsection{Data Diversity}
\begin{figure*}[!t]
\centering
\setkeys{Gin}{width=0.14\linewidth}
\subfloat{\includegraphics{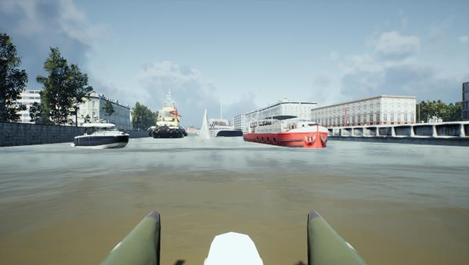}}
\hfill
\subfloat{\includegraphics{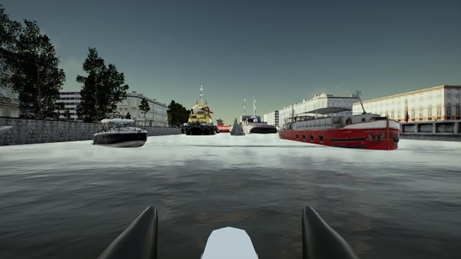}}
\hfill
\subfloat{\includegraphics{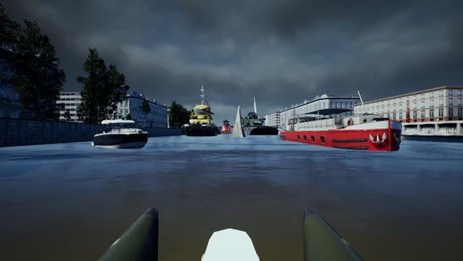}}
\hfill
\subfloat{\includegraphics{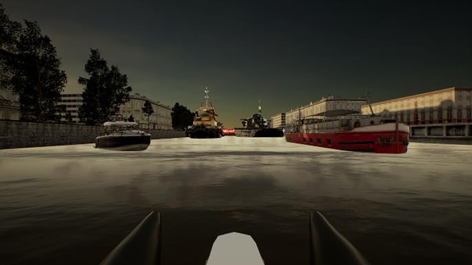}}
\hfill
\subfloat{\includegraphics{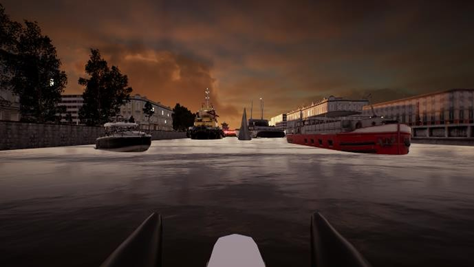}}
\hfill
\subfloat{\includegraphics{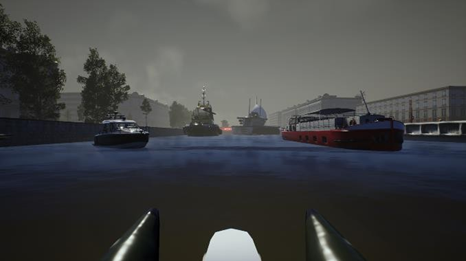}}
\hfill
\subfloat{\includegraphics{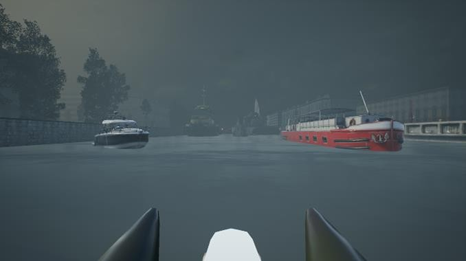}}
\hfill
\caption{Weather and Illumination variations in SimuShips}
\label{fig:weather_and_illumination}
\end{figure*}

\begin{figure*}[!t]
\centering
\setkeys{Gin}{width=0.2\linewidth}
\subfloat{\includegraphics{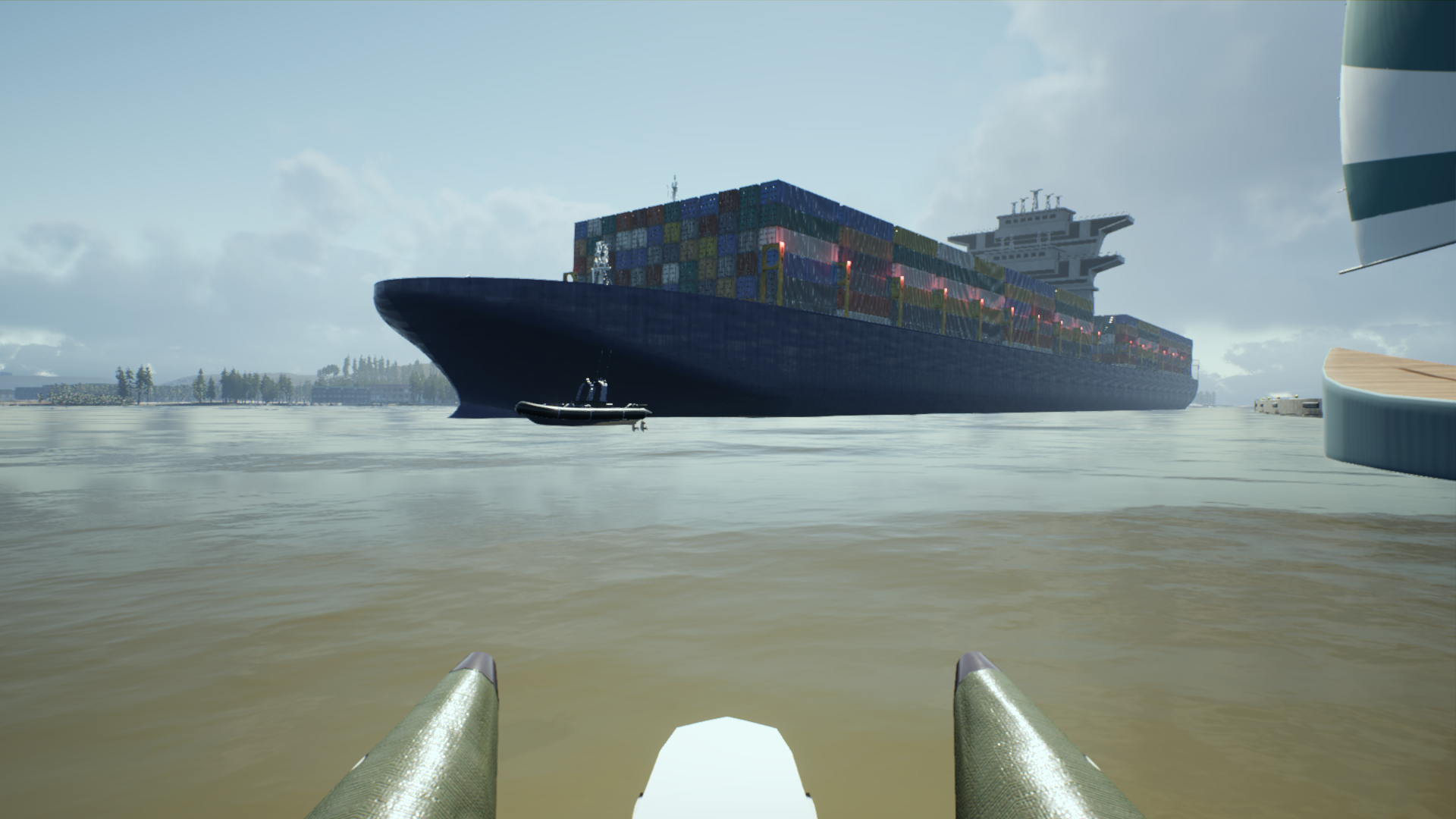}}
\hfill
\subfloat{\includegraphics{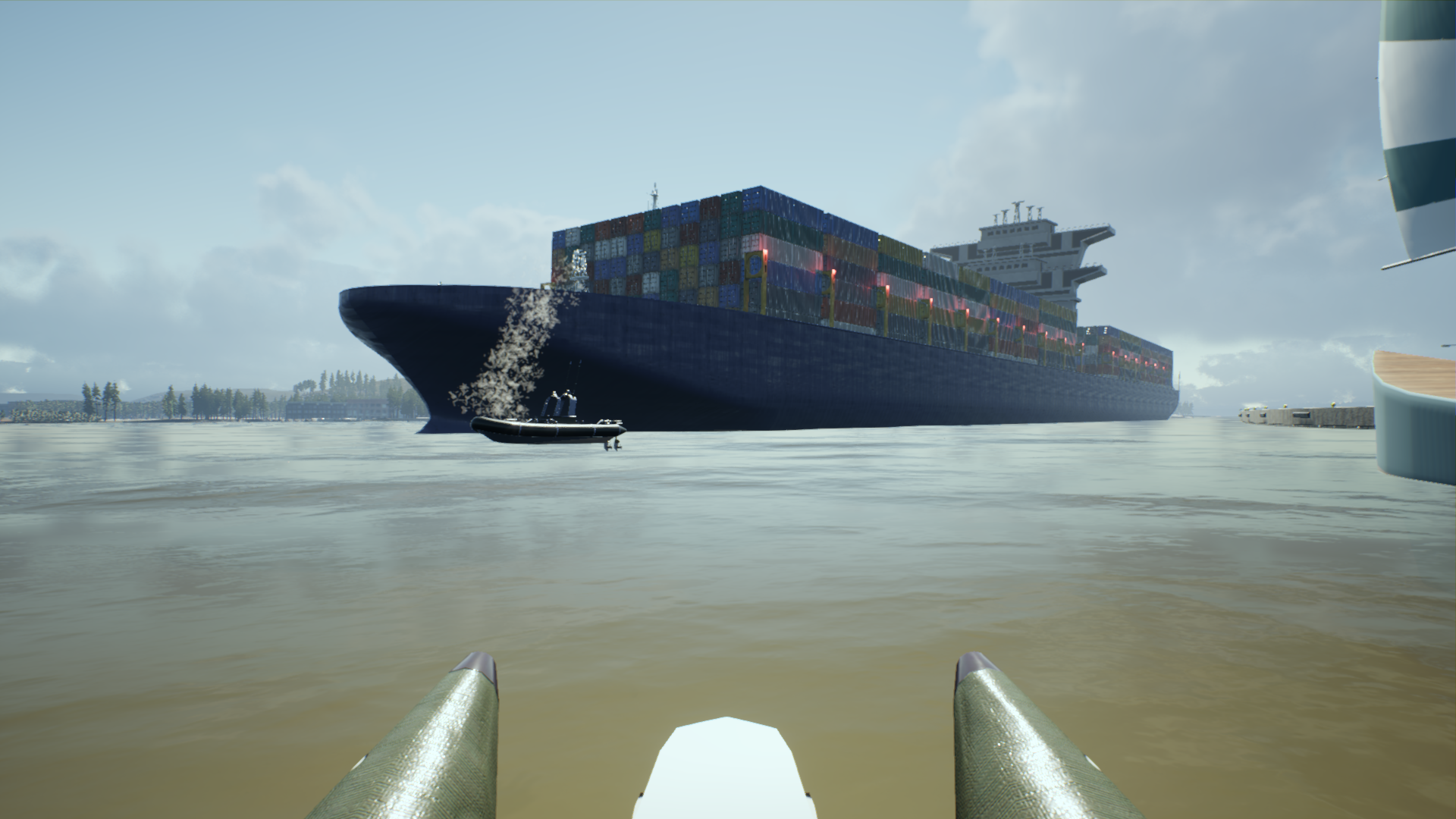}}
\hfill
\subfloat{\includegraphics{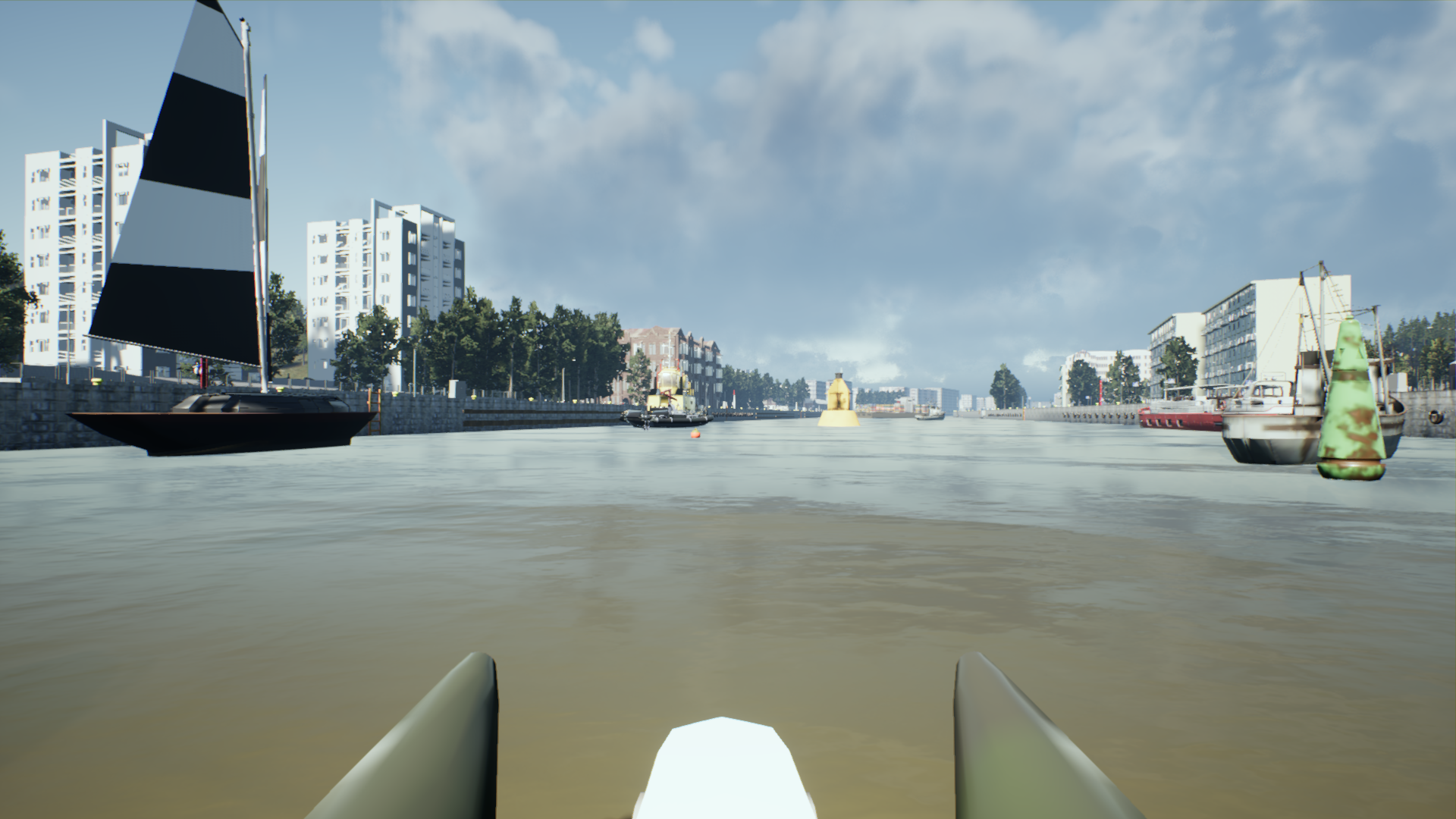}}
\hfill
\subfloat{\includegraphics{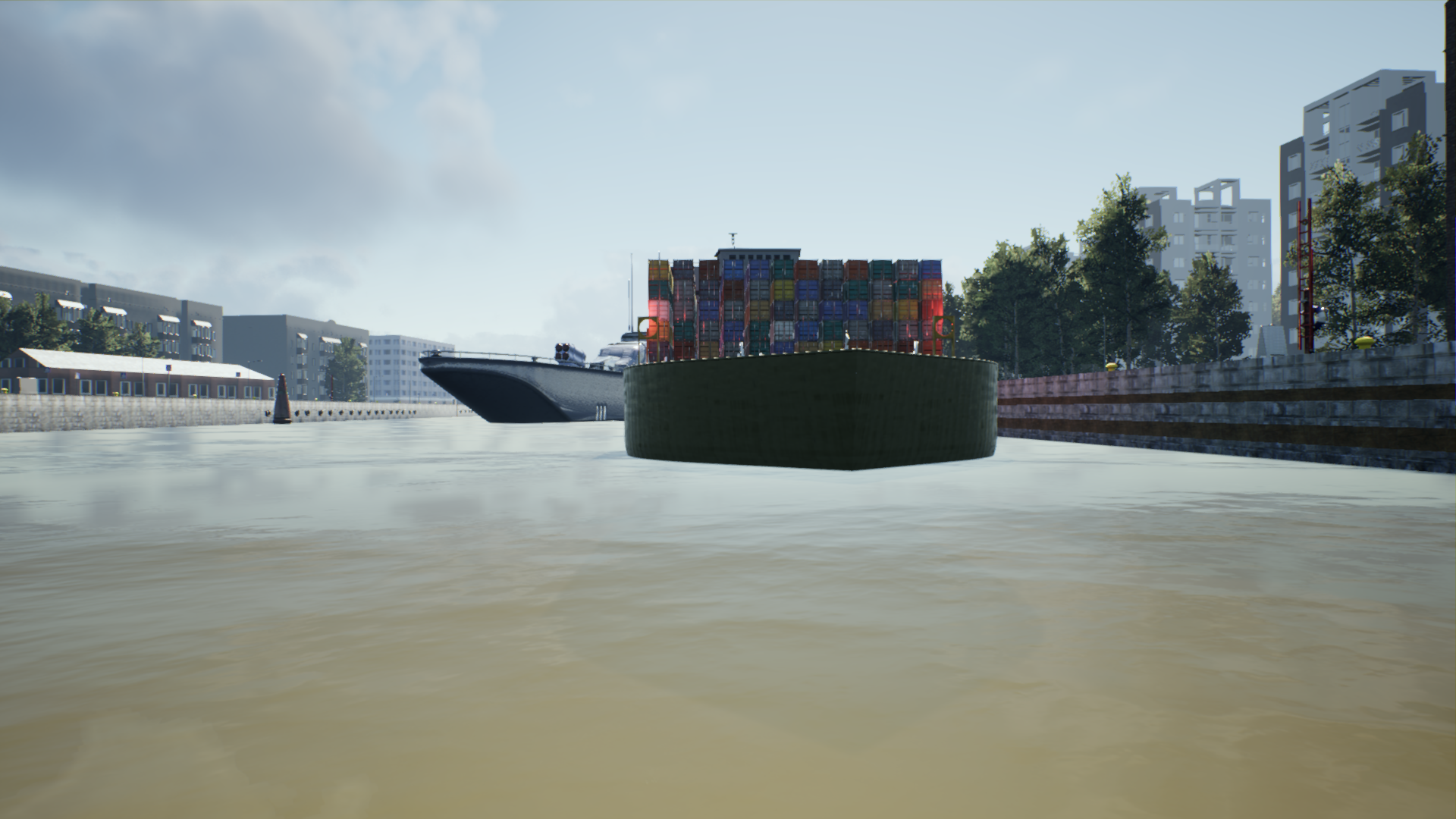}}
\hfill
\subfloat{\includegraphics{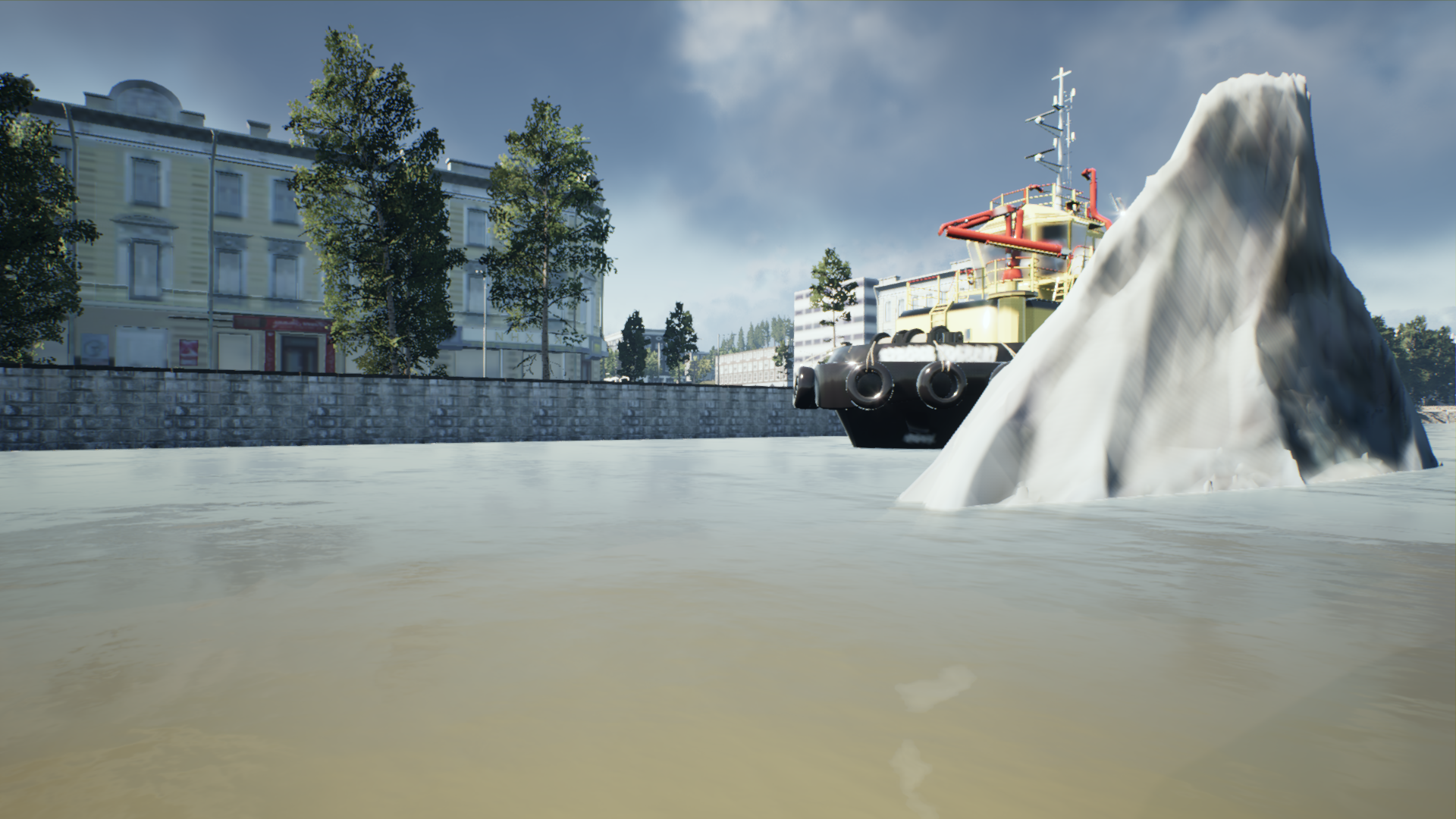}}
\hfill
\subfloat{\includegraphics{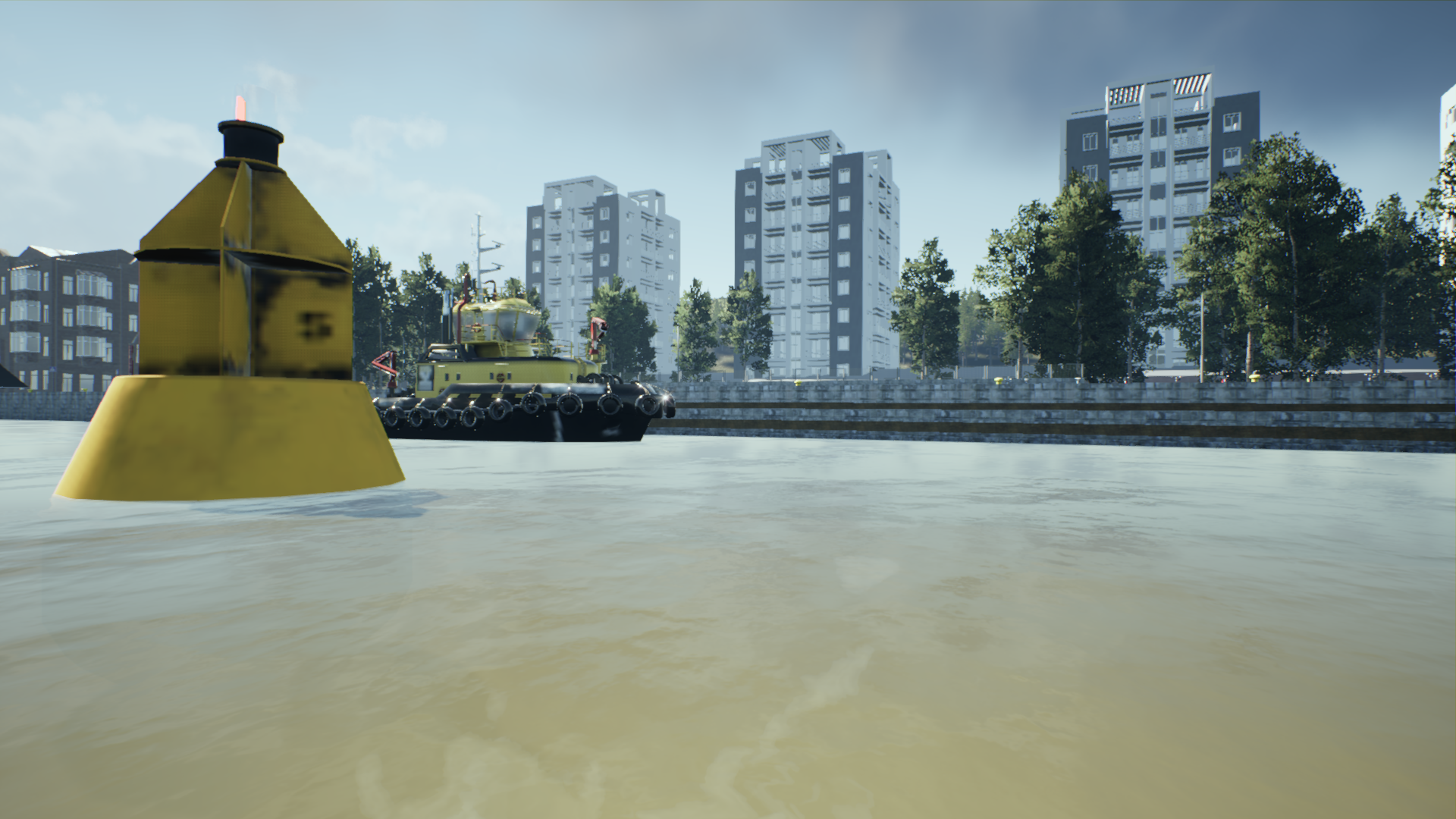}}
\hfill
\subfloat{\includegraphics{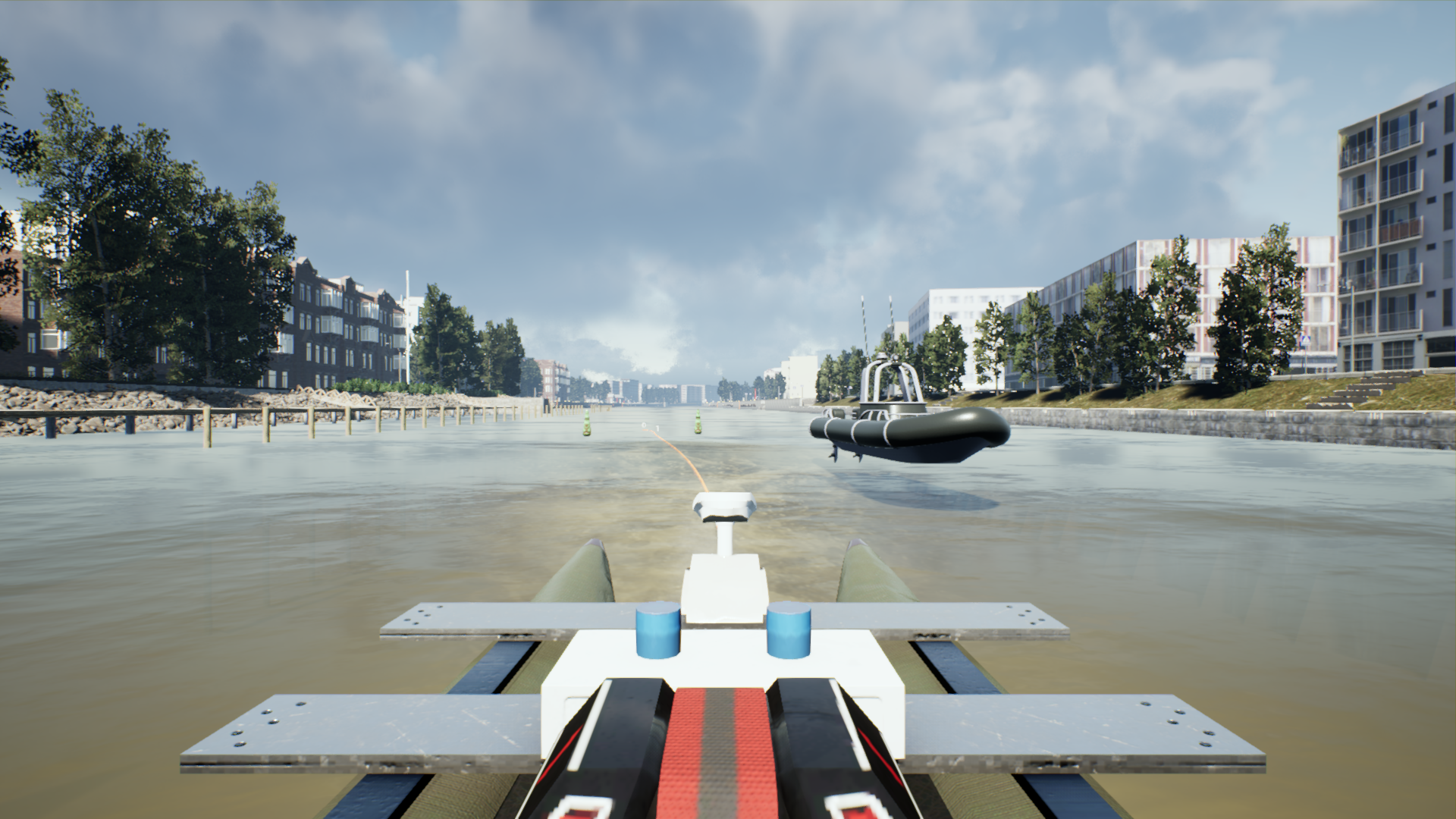}}
\hfill
\subfloat{\includegraphics{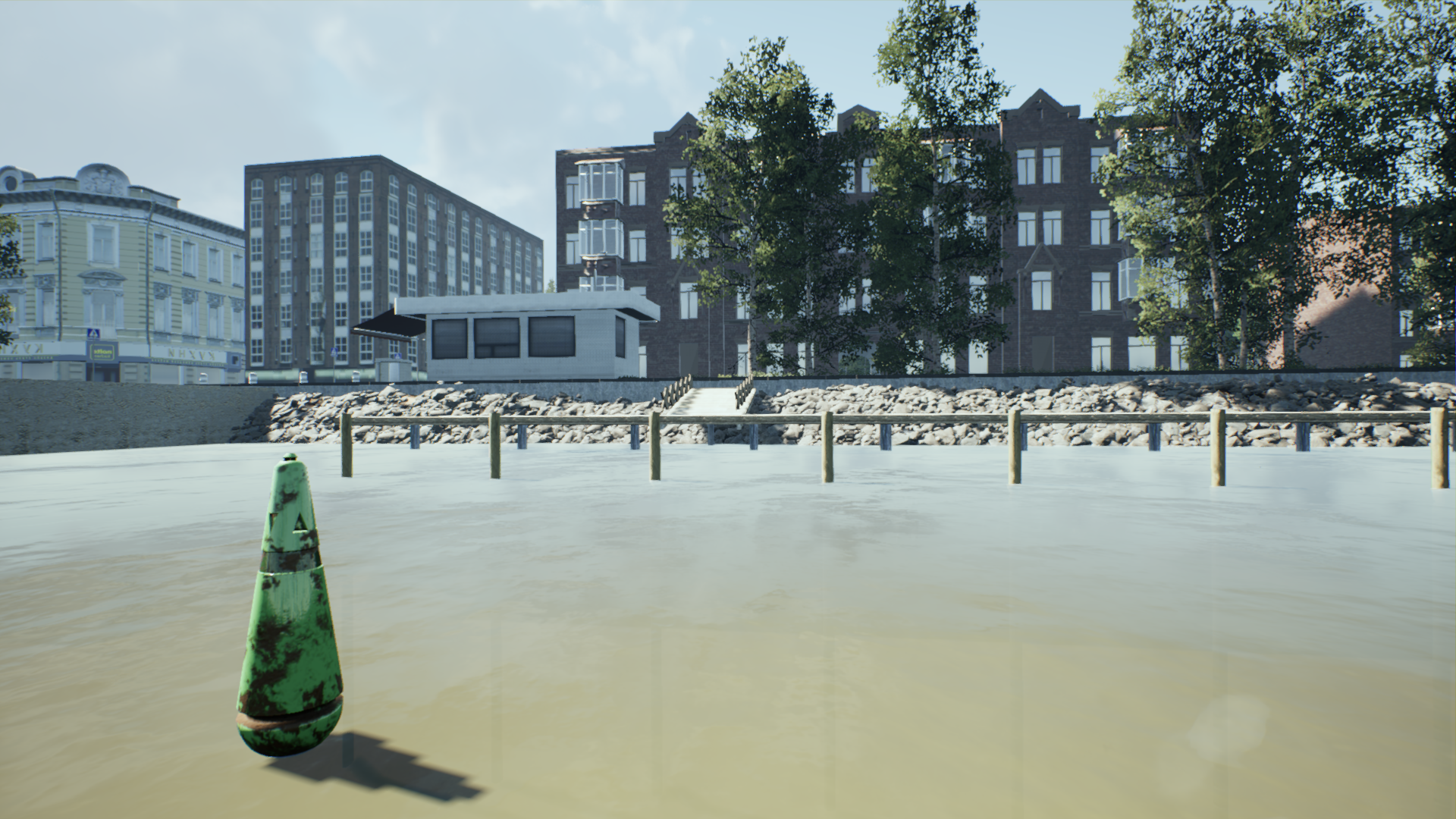}}
\hfill
\subfloat{\includegraphics{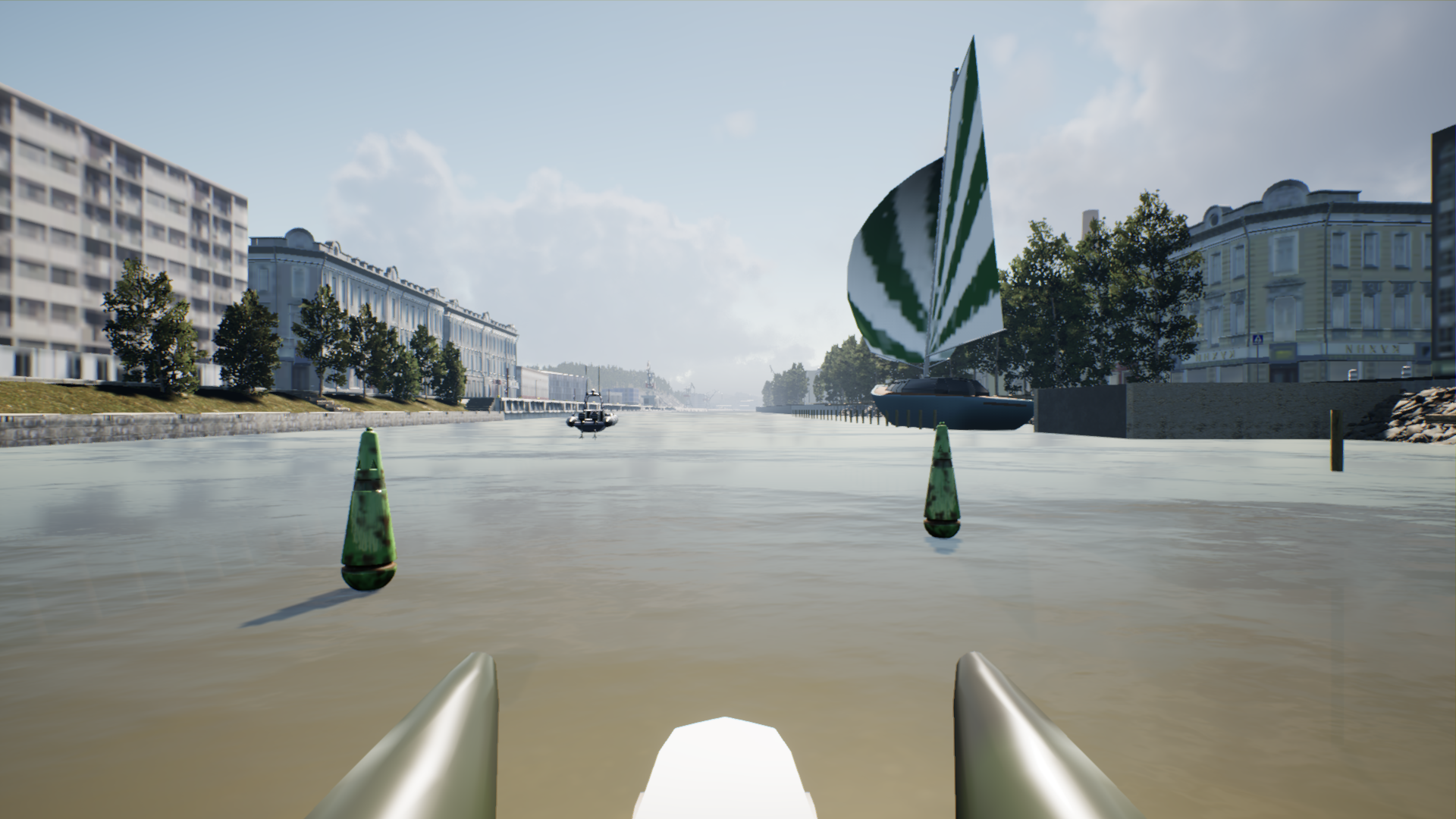}}
\hfill
\subfloat{\includegraphics{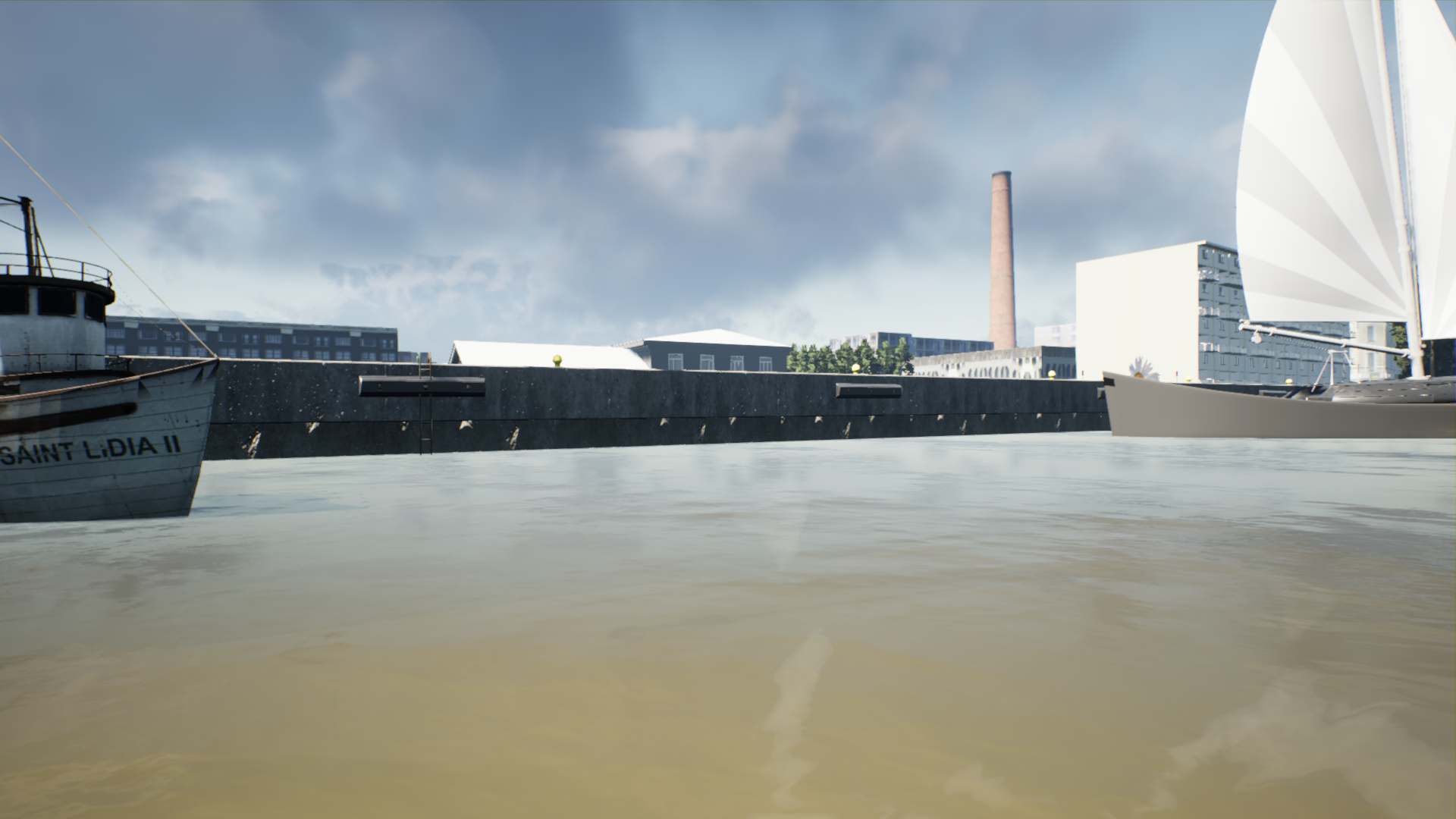}}
\caption{ Objects vary in scale and visible proportions. The dataset also takes occlusion into consideration and incorporates scenarios where one object covers the other.}
\label{fig:data_diversity}
\end{figure*}

While collecting the images, the complex and intricate nature of maritime environments was taken into consideration. In particular, we focused on including a wide range of obstacle types, atmospheric and illumination conditions along with occlusion, scale and proportion variations which can be seen in Fig. \ref{fig:weather_and_illumination} and Fig. \ref{fig:data_diversity}.

\emph{Obstacles types}: Our dataset includes a variety of maritime vessels diverging in size and configuration. The vessel types include motorboats, tugboats, warships, yachts, sailboats (with different sail variants), fishing boats, passenger ships(ferries) and cargo ships. We include static obstacles as well (rock and buoy) with size and colour variations.

\emph{Atmospheric conditions}: Commonly occurring weather conditions were considered including clear, cloudy and foggy.

\emph{Illumination conditions}: Since light variations can significantly impact the captured images, we considered illumination conditions at different times of the day including afternoon, evening and night.

\emph{Visible proportions}: Our dataset includes images with different visible proportions of the objects. Even when the object is slightly visible in the image, a label is assigned to train a more robust model.

\emph{Occlusion}: Since the given dataset aims to accurately depict physical context, we include scenarios where maritime vessels occlude each other and/or other objects in the background.

\emph{Scale variation}: Maritime environments include ships across a range of distances which affects their size in the captured images. A suitable dataset for object detection captures large as well as small objects in the image. Therefore, we focus on capturing vessels at multiple distances.

\subsection{Data Annotation}
\begin{figure}[!t]
\centering
\setkeys{Gin}{width=0.3333\linewidth}
\subfloat{\includegraphics{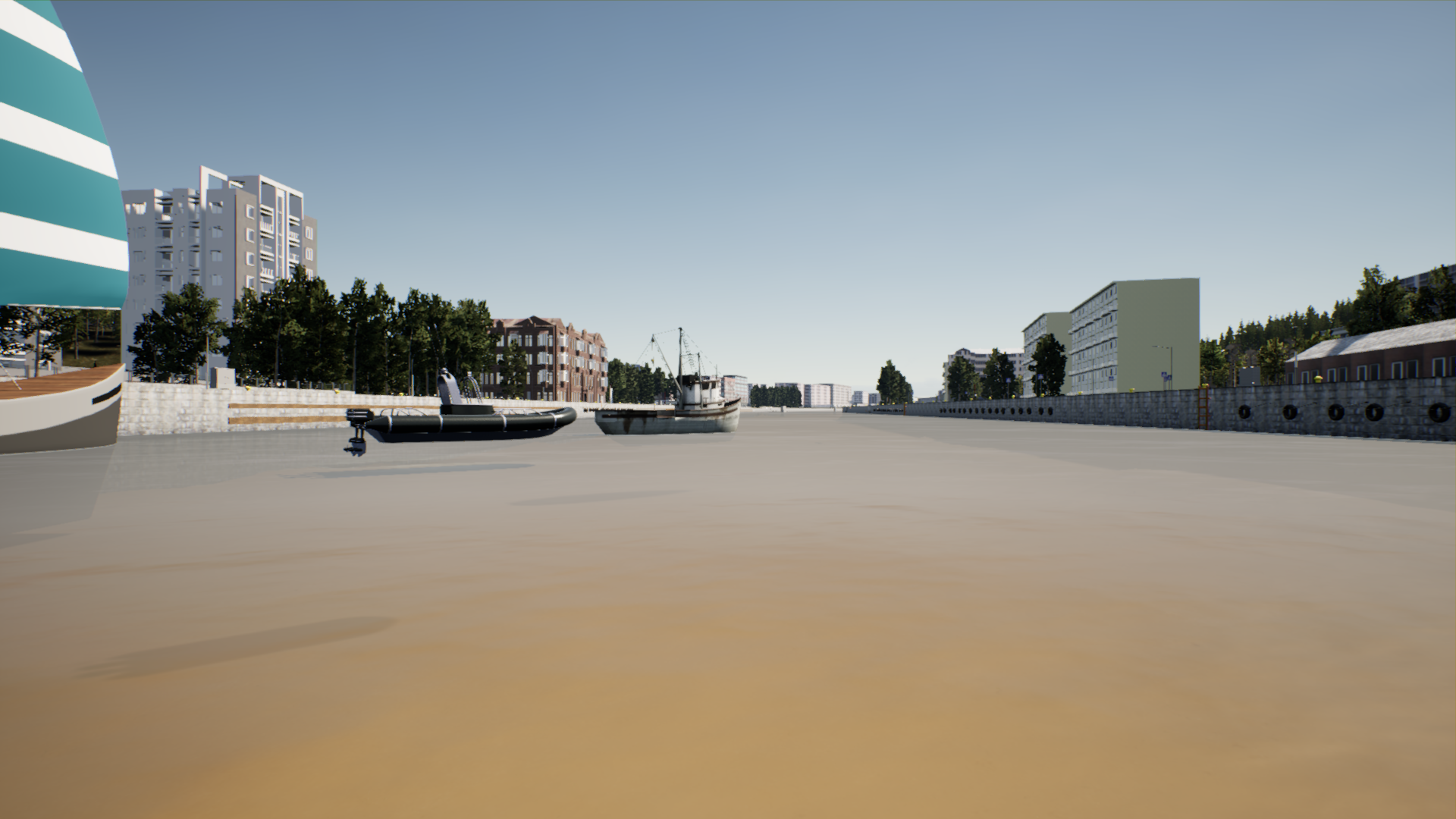}}
\hfill
\subfloat{\includegraphics{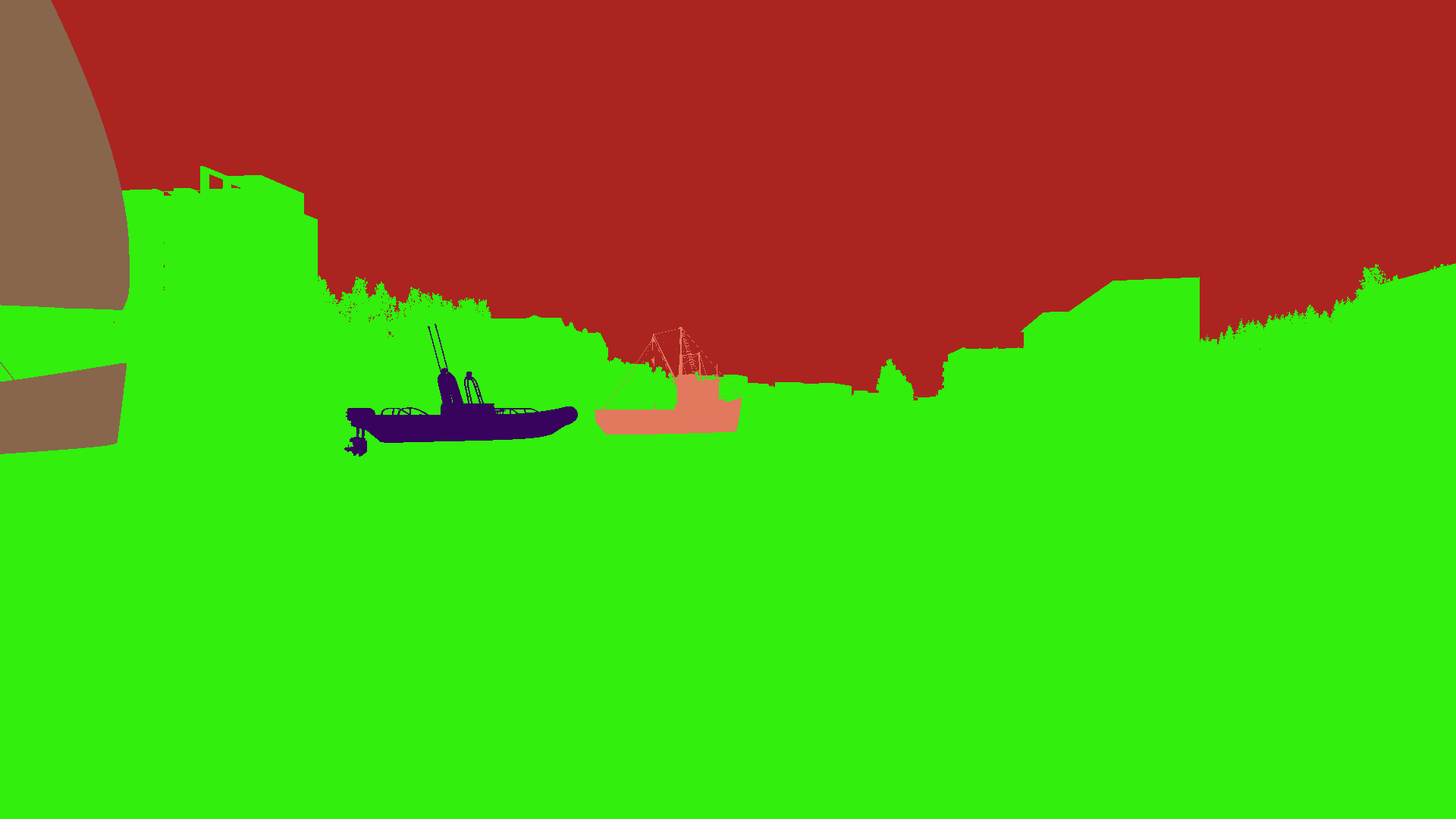}}
\hfill
\subfloat{\includegraphics{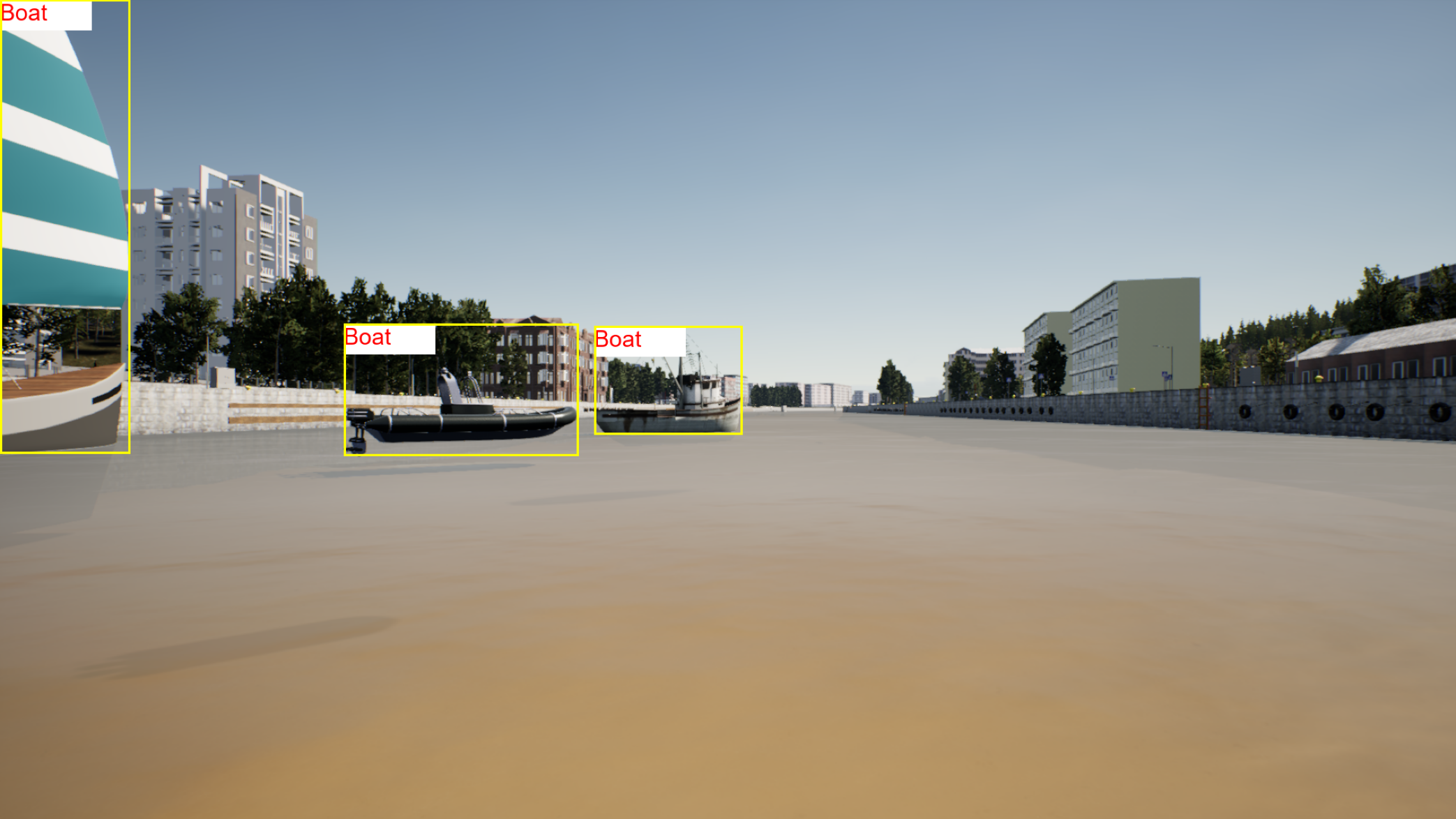}}
\caption{Annotation pipeline showing the RGB image, segmentation mask and the annotated image}
\label{fig:annotation}
\end{figure}

After collecting the RGB images and their corresponding segmentation masks from the simulation tool, the annotations were extracted using an open-source Python library, OpenCV. The object annotation pipeline is shown in Fig. \ref{fig:annotation}. The colors in the segmentation masks were mapped to categories. For each unique color in the segmentation mask, a mask was created which contains the occurrences of that color only. The image was converted to a binary image by applying a threshold. The contours (edges where color changes sharply) were extracted from the binary image which were finally used to find the bounding boxes. The output is in the form of a tuple $(o, x, y, w, h)$ where $(x,y)$ is the top left corner of the bounding box, $w$ and $h$  are the width and height of the bounding box respectively and $o$ is the object category. We convert this into the format used by the YOLOv5 model. Similar to the $(o, x, y, w, h)$ format, each bounding box is represented by five values. These are the center of the box, its width and height and the object category. One major difference is that these values are normalized by the dimensions of the image (1920x1080). We provide a text file for each image which contains the annotations separated by a new line.

We divided the annotations into two categories.
\begin{itemize}
    \item Dynamic Obstacles: This category includes all maritime vessels.
    \item Static Obstacles: This category includes rocks and buoys.
\end{itemize}
The intuition behind the choice of these categories was that maritime vessels may have dynamic properties. Therefore, to avoid the vessels, their dynamics must be considered which can be captured via other sensors. On the other hand, buoys and rocks are stationary. The specifications of the dataset are summarized in Table \ref{tab1}
\begin{table}[htbp]
\caption{Characteristics of SimuShips Dataset}
\begin{center}
\begin{tabular}{|l|c|}
\hline
Total number of images                                & 6756 + 2715 = 9471 \\
Images (without weather augmentation) & 2715\\
Range of number of objects in images                  & [1, 10] 
\\
Categories                                            & Dynamic (0), Static (1) \\           
\hline
\end{tabular}
\label{tab1}
\end{center}
\end{table}


\section{Experiments and Results} \label{exp}
We comprehensively evaluated the efficacy of augmenting real and simulated data. We use a publicly available dataset, ABOships \cite{iancu} which has been collected in Turku, Finland. We use YOLOv5 ('You only look once') which is a family of end-to-end object detection algorithms. These lightweight models process images at a very high speed without compromising the accuracy of inference. Yolov5 uses R-CNN (Region Proposal + CNNs) to compose a solution from the grid, each part of which detects objects within its own area \cite{yolov5docs}. Prior experimental results indicate that YOLOv5, latest version of YOLO, has the highest speed (up to 140 fps) and smaller size than the previous models \cite{yolov5_drones, yolov5_fish}. It exceeds Faster R-CNN \cite{faster_rcnn} with regards to precision and recall rates making it suitable for AMSVs. Before discussing the results, we define the metrics used to compare performance.
\begin{table*}[t]
\caption{Results of YOLOv5 Object Detection}
\begin{center}
\begin{tabular}{|c|c|c|c|c|c|c|c|c|}
\hline
Test Set & Model  & Images&  Class & Labels & Precision & Recall & mAP@.5 & mAP@.5:.95 \\
\hline
& &  & & & & & &\\
  &   &  & all & 4504 & 0.522 & 0.271 & 0.281 & 0.162\\
  & A &  & boat & 3967 & 0.574 & 0.343 & 0.37 & 0.194 \\
  &   & &obstacle & 537 & 0.47 & 0.199 & 0.192 & 0.131\\
SimuShips& & 1894 & & & & & &\\
  &   &  & all & 4504 & 0.958 & 0.837 & 0.873 & 0.714\\
  & B & & boat & 3967 & 0.965 & 0.858 & 0.904 & 0.773 \\
  &   & & obstacle & 537 & 0.95 & 0.817 &0.841 & 0.655\\
\hline
  &   &  & all & 8362 & 0.698 &0.561 &      0.614 &     0.25\\
  & A &  & boat & 6776 & 0.703 &     0.653 &     0.70 &     0.32  \\
  &   & &obstacle & 1586 &0.692 &     0.468 &     0.524 &     0.177\\
ABOShips& & 1807 & & & & & &\\
  &   &  & all & 8362 & 0.674  &    0.577 &     0.618 &0.2505\\
  & B & & boat & 1586 & 0.665   &   0.667  &    0.695  &     0.32  \\
  &   & & obstacle & 537 & 0.683 &     0.486    &   0.54&       0.18  \\ 
\hline
\end{tabular}
\label{tab_results}
\end{center}
\end{table*}
\subsection{Evaluation Metrics}
Various quantitative indicators can be used to evaluate the performance of object detection models. A popular measure is Intersection Over Union (IoU). IoU is a measure based on Jaccard Index which evaluates the extent of overlap between two bounding boxes – the ground truth (actual) and the prediction. IoU is given by the overlapping area between the predicted bounding box $B_p$ and the ground truth bounding box $B_{gt}$ divided by the area of union between them \cite{metrics} as shown in (\ref{eq:Iou}):
\begin{equation}\label{eq:Iou}
    IoU = \frac{B_p \cap B_{gt}}{B_p \cup B_{gt}}
\end{equation}
A threshold value $thres$ for IoU determines whether the box belongs to the background $IoU < thres$ or is counted as a prediction $IoU \geq thres$. With this metric, recall and precision can be defined. 
\emph{Recall} reflects an algorithm's capability to capture objects from the image i.e. out of all the predictions present in the dataset, how many were successfully predicted by the model. Recall is a crucial metric for AMSVs since missing obstacles can result in accidents. \emph{Precision} reflects an algorithm's capability to identify relevant objects. It is calculated as the proportion of correctly identified bounding boxes over the total number of detected boxes. 
\emph{F1 score} is the harmonic mean of precision and recall which serves as a single numerical metric to compare the performance of models. \emph{Average precision (AP)} is another metric derived from precision and recall. It is the area under the precision-recall curve. The mean average precision (mAP) is calculated by finding AP for each class and then taking an average over the number of classes. Depending upon the IoU threshold, we have used two metrics based on mAP. mAP@0.5 is the mAP value for IoU threshold 0.5 while mAP@0.5:0.95 is the mAP average over different IoU thresholds, from 0.5 to 0.95.

\subsection{Quantitative Results}
In this work, we used the open-source implementation by Ultralytics. First, we randomly split the two datasets into training and test sets in a 80/20 ratio. Then, we merged and shuffled the training datasets to get a combined dataset. We trained YOLOv5 on ABOships (Model A). We trained the same architecture on the combined dataset (Model B). We evaluated these models on SimuShips and ABOShips (see Table \ref{tab_results}). Model B outperformed Model A on both test sets. In particular, for the real images, Model B achieved a higher recall as well as mAP score. 

\begin{figure}[!t]
\centering
\includegraphics[width = 3.5in]{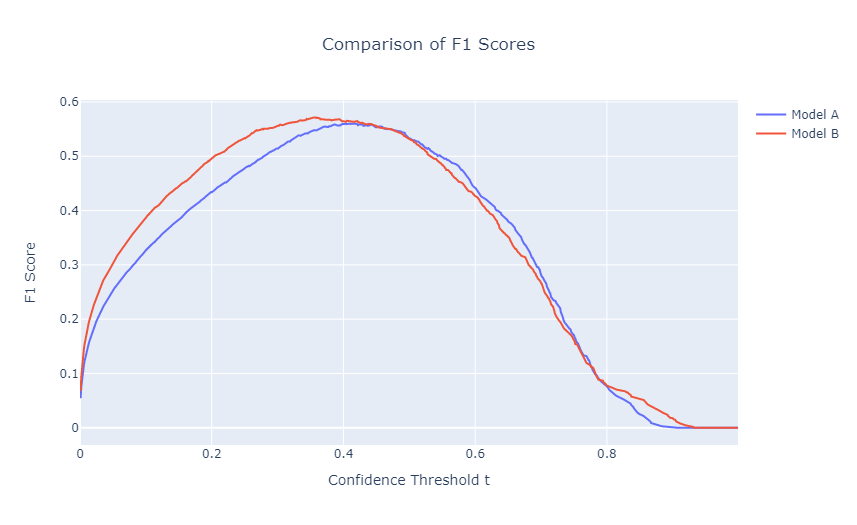}
\caption{F1 curves for Model A and B indicating that Model B achieves a higher area under the curve}
\label{fig:f1curve}
\end{figure}

Since precision and recall are sensitive to the confidence threshold, we used F1 curves across a range of confidence thresholds (1000 points) to compare the performance of the two models. As indicated by Fig. \ref{fig:f1curve}, around default values of the confidence thresholds, Model B had a higher F1 score as compared to Model A. Model B achieved a higher area under the curve as well.

\section{Conclusion} \label{conclusion}
In this paper, we present SimuShips, a high-resolution, simulation-based dataset for maritime environments consisting of images resembling the real world. Our dataset incorporates diversity of objects, weather, illumination, visible proportion and scale. Moreover, we conduct experiments with our dataset and a publicly available dataset, ABOships, containing only real world images, to evaluate the performance of YOLOv5.  We observed that the combination of real and simulated images improves the recall for all classes by 2.9\%. These experiments highlight a novel direction for researchers to collect data from digital tools and combine it with the real world data to solve machine learning and deep learning problems. We consider it as an open research avenue to augment image data with domain-specific simulation data for other computer vision tasks such as semantic segmentation, classification and Simultaneous Localization and Mapping (SLAM) and analyze the effectiveness of such amalgamation.

\section*{Acknowledgment}
The work has been partially supported by the EMJMD master’s programme in Engineering of Data-Intensive Intelligent Software Systems (EDISS - European Union’s Education, Audiovisual and Culture Executive Agency grant number 619819). We would like to express our gratitude to Jérôme Leudet at AILiveSim Oy for his help and technical support with the 3D virtual development platform.

\bibliographystyle{IEEEtran}
\bibliography{references}

\end{document}